\documentclass[a4paper]{article}
\usepackage{multirow,multicol, makecell}
\usepackage{ISCSLP2022}
\usepackage{hyperref}
\usepackage{graphicx}
\usepackage{subfigure} 
\title{Rhythm-controllable Attention with High Robustness for Long Sentence Speech Synthesis}
\name{Dengfeng Ke$^{1,\dagger}$, Yayue Deng$^{2,1,\dagger}$, Yukang Jia$^{1,\dagger}$, Jinlong Xue$^{2,\dagger}$, Qi Luo$^1$, Ya Li$^{2,*}$, Jianqing Sun$^3$, Jiaen Liang$^3$, Binghuai Lin$^4$} 

\address{
  $^1$ Beijing Language and Culture University, Beijing, China\\
  $^2$ School of Artificial Intelligence, Beijing University of Posts and Telecommunications, Beijing, China\\
  $^3$ Unisound AI Technology Co., Ltd, Beijing, China
  \\
  $^4$ Smart Platform Product Department, Tencent Technology Co., Ltd, China}
\email{dengfeng.ke@blcu.edu.cn, dengyayue.stu@gmail.com, marshal\_jia@foxmail.com, jinlong\_xue@bupt.edu.cn, lqttkxljys@163.com, yli01@bupt.edu.cn, sunjianqing@unisound.com, liangjiaen@unisound.com, binghuailin@tencent.com
}
\begin{document}

\maketitle
\renewcommand{\thefootnote}{}
% \footnotetext{\dagger \; Equal \: contribution. * \; Corresponding \: author}
\footnotetext{$\dagger$ Equal contribution. $*$ Corresponding author}

\begin{abstract}
  Regressive Text-to-Speech (TTS) system utilizes attention mechanism to generate alignment between text and acoustic feature sequence. Alignment determines synthesis robustness (e.g, the occurence of skipping, repeating, and collapse) and rhythm via duration control. However, current attention algorithms used in speech synthesis cannot control rhythm using external duration information to generate natural speech while ensuring robustness. In this study, we propose Rhythm-controllable Attention (RC-Attention) based on Tracotron2, which improves robustness and naturalness simultaneously. Proposed attention adopts a trainable scalar learned from four kinds of information to achieve rhythm control, which makes rhythm control more robust and natural, even when synthesized sentences are extremely longer than training corpus. We use word errors counting and AB preference test to measure robustness of proposed
method and naturalness of synthesized speech, respectively. Results shows that RC-Attention has the lowest word error rate of nearly 0.6\%, compared with 11.8\% for baseline system. Moreover, nearly 60\% subjects prefer to the speech synthesized with RC-Attention to that with Forward Attention, because the former has more natural rhythm.
  
\end{abstract}

\begin{figure*}[t]
  \centering
  \includegraphics[width=\linewidth]{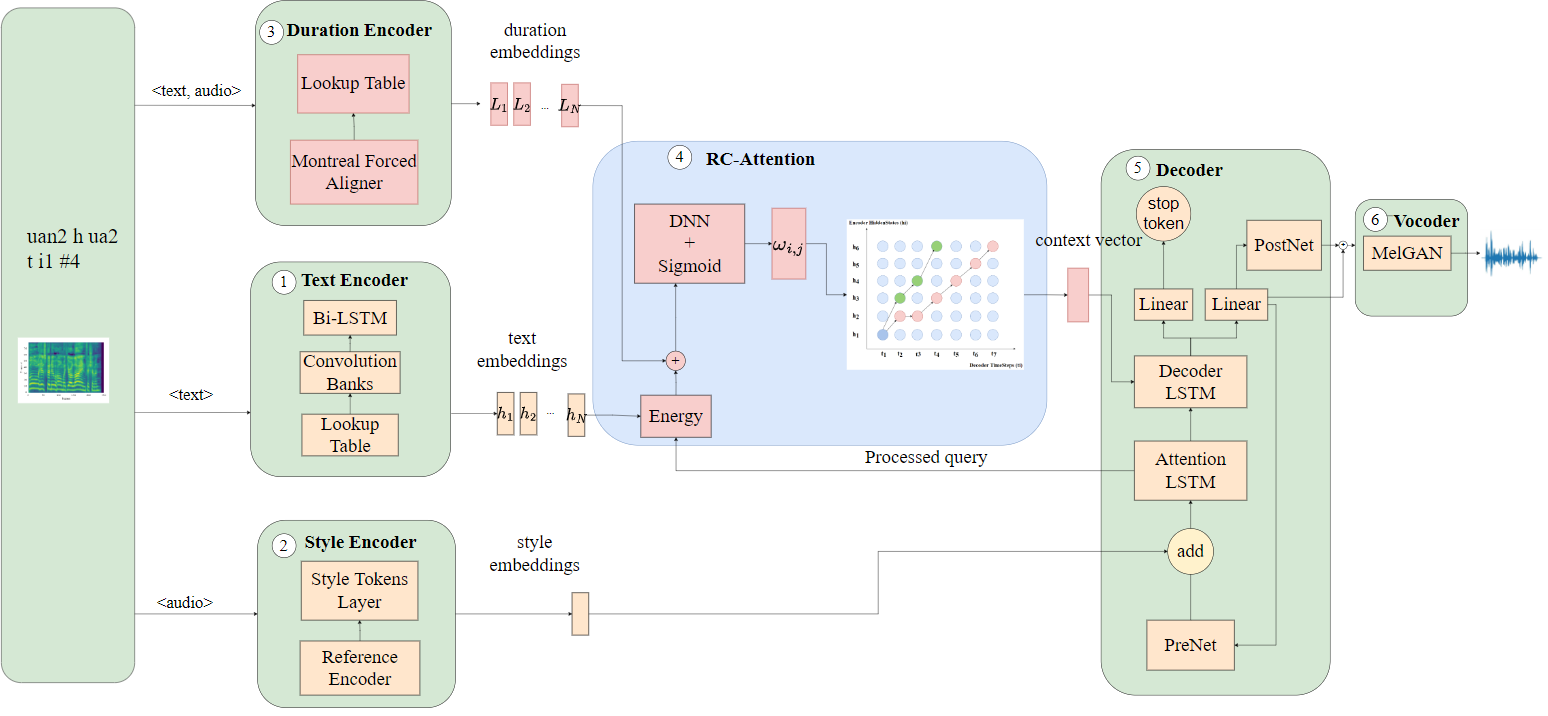}
  \caption{The architecture of the whole speech synthesis model. Blue box shows computational procedure of RC-Attention. ‘$\omega_{i,j}$’, ‘$+$’ and ‘add’ represents a trainable scalar, concatenation and addition operation, respectively.}
  \label{fig:whole model}
\end{figure*}

\section{Introduction}

Current end-to-end speech synthesis systems \cite{oord2016wavenet,ren2020fastspeech,chen2021adaspeech,kim2020glow,miao2020flow} have the ability to generate high-quality and human-like speech. Hence, speech synthesis application scenarios are becoming more and more diversified. All of these scenarios have two basic demands on synthesis systems.
Firstly, speech corresponding to text can be synthesized accurately and does not exist repetition, skipping, or gibberish. This demand focuses on the robustness of synthesis system. Secondly, in addition to accuracy, people would expect synthesized speech is capable of generating natural rhythm of human. This demand focuses on model's ability of modeling and controlling of rhythm. Both of demands can be satisfied by a well-designed attention mechanism.

The problems of robustness in speech synthesis arise from extremely unequal length between text sequence and acoustic feature sequence. In the process of upsampling, if mapping relationship from text sequences to acoustic feature sequences, named alignment matrix, is not learned well, two types of alignment problems are likely to occur. One is attention collapse which will lead to gibberish. The other is blurred  alignment which means acoustic model fails to focus on a single input token in a decoding step, leading to skipping and repeating.

Some studies \cite{chorowski2015attention,tian2020feathertts,chen2020multispeech,chiu2017monotonic} propose robust attention mechanisms in order to alleviate difficulty of alignment learning and accumulation of errors in the inference stage. Raffel et al. \cite{raffel2017online} propose Monotonic Attention. The core idea is that when decoding the current timestep, model only needs to decide whether to focus on the current phoneme or move the focus forward. However, completeness of alignment cannot be satisfied, which may lead to skipping. He et al. \cite{he2019robust} propose Stepwise Monotonic Attention based on Monotonic Attention. By adding a constraint of forward stride to monotonic attention, the method ensures that each phoneme can be covered by at least one frame of mel spectrogram. Another novel attention is propoesd by Graves et al. \cite{graves2013generating}, named GMM Attention which uses multiple mixed Gaussian distributions to model alignment. Meanwhile, in order to make sure monotonicity, the mean value of mixed Gaussian is constrained to increase as decoding time goes by. Many speech synthesis systems use these attentions to speed up converge of training and improve robustness of synthesis \cite{liu2021reinforcement,yang2022improving}.

Methods of rhythm control are investigated by a few studies. For example, Zhang et al. \cite{zhang2018forward} propose a forward algorithm, named Forward Attention which contains a transition agent to achieve rhythm control. However, robustness may be damaged after adding the external rhythm control agent. In the inferring phase, adjusting value of transition agent by human may leads to accumulation of errors.

Therefore, in this work, we combine advantages of these attentions and propose a novel attention mechanism, called Rhythm-controllable Attention (RC-Attention), to satisfy demands of robustness and rhythm control. Our attention mechanism can improve robustness compared with other advanced attention mechanisms, especially in long sentences synthesis, meanwhile utilizing external duration information to synthesize speech with more natural rhythm.

\section{Attention-based neural text-to-speech}
\label{related_work}
Given an input sequence $X = [x_1, x_2, … , x_{N}]$ with length N, text encoder produces a sequence of hidden state $H=[h_1, h_2, … , h_N]$ which is convenient for attention mechanism to use, making the training more stable.

The mathematical representation of the first recurrent neural structure in the attention-based TTS is illustrated as follow:
\begin{equation}
s_i=LSTM\_Cell_{att}(s_{i-1},c_{i-1},y_{i-1})
\label{equ2}
\end{equation}
where $c_{i-1}$ and $y_{i-1}$ represent context vector and acoustic feature of previous decoding step, respectively. They are concatenated and sent into attention LSTM cell, producing current LSTM state $s_{i}$.

The general form of attention is shown in Eq.~(\ref{equ3}), where $e_{i,j}$ represents energy value of the $j$-th phoneme at the $i$-th decoding step.
\begin{equation}
e_{i,j}=Attention(s_i,\,h_j,\,…)
\label{equ3}
\end{equation}

\begin{equation}
a_{i,j}=Softmax(e_{i,j})
\label{equ4}
\end{equation}

After alignment vector $a_{i,j}$ is calculated as shown in Eq.~(\ref{equ4}), context vector $c_i$ is obtained by weighted sum of the hidden state sequences $H$.
\begin{equation}
c_i=\sum_{j=1}^{N} a_{i,j}{h_{j}}
\label{equ5}
\end{equation}

\section{Proposed method}
\label{model}
Previous attention mechanisms or alignment tricks mostly focus on improving robustness of synthesis. Few alignment algorithms can achieve rhythm control by utilizing external duration information. Even if they can control rhythm, robustness would be damaged after adding rhythm controller. For example, after using Forward Attention with a transition agent, robust problems of synthesized speech would appear more frequently. Inspired by this, we integrate advantages of previous attention mechanisms and propose RC-Attention to achieve rhythm control in phoneme level without destroying robustness.
 
\subsection{The whole speech synthesis system architecture}
The overall text-to-speech system is illustrated in Fig.~\ref{fig:whole model}. 
Our model uses Tacotron2 \cite{shen2018natural} as backbone. The whole synthesis model contains six parts: 1) text encoder to embed input sequence into hidden state sequence, 2) style encoder to extract style features from input audio, 3) duration encoder to map duration information to a fixed vector, 4) RC-Attention to align hidden state sequence with acoustic feature sequence, 5) decoder to decode current acoustic feature frame and 6) vocoder to synthesize waveform of speech from acoustic feature. 

In the training phase, \emph{(text, audio)} pairs are sent into the model. Text encoder takes phoneme id sequence as input to extract text hidden state sequence, namely text embeddings in Fig.~\ref{fig:whole model}. Style encoder structure is the same as global style tokens \cite{wang2018style}, generating style embedding from mel spectrograms. Inspired by \cite{lee2017emotional}, we inject style embedding into the attention LSTM by adding it to the output of PreNet. Duration encoder takes \emph{(text, audio)} pairs as input and utilizes forced alignment tool to get duration of each phoneme. The duration sequence is mapped into duration embeddings by lookup table. In the inferring phase, duration sequence can be externally specified to achieve external control of rhythm. The addition of style embedding and previous acoustic frame processed by PreNet. The result is fed into the first recurrent neural structure as shown in Eq.~(\ref{equ2}). The energy function adopts additive attention mechanism which takes current state of Attention LSTM and text embeddings as query and key, respectively. RC-Attention takes the concatenation of energy value and duration embeddings as input and calculates context vector. The calculative process will be illustrated in the Section~\ref{rc-attention}. 

\subsection{Rhythm-controllable attention}
\label{rc-attention}
\begin{figure}[htp]
  \centering
  \includegraphics[width=160pt,height=110pt]{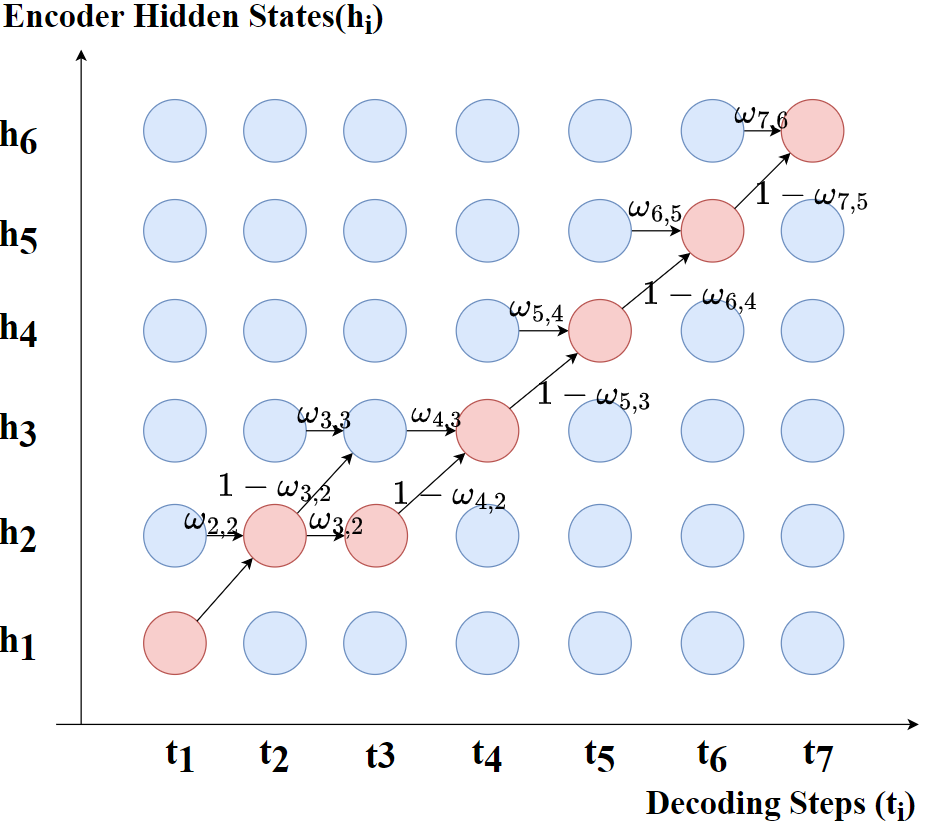}
  \caption{Calculation process of alignment in RC-Attention. Vertical axis represents encoder hidden state sequence generated by text encoder. Horizontal axis represents decoding step.}
  \label{fig:ali_visual}
\end{figure}

 RC-Attention is proposed based on two assumptions which guarantee robustness and natural rhythm of synthesized speech. 
 Inspired by Forward Attention \cite{zhang2018forward}, the first assumption is illustrated as follow: The text encoder hidden state noticed by the current decoding step can only be the encoder hidden state noticed by the previous decoding step or next one. 
 As shown in the Fig.~\ref{fig:ali_visual}, there are only seven decoding steps. Take the decoding step $t_4$ as example, index of max value in the alignment vector at decoding step $t_4$ can only be the index of max value in decoding step $t_3$ or its following one, which is $h_2$ or $h_3$ respectively. Besides, each column only has a maximum value, which means at this decoding step acoustic model would generate acoustic feature frame according to the respective encoder hidden state. Hence, the first assumption guarantees monotonicity and completeness of alignment path.

 The second assumption utilizes a trainable scalar $\omega$ which learns from four kinds of information: acoustic information, text information, style information and duration information. The text information refers to the hidden state sequences of encoder. Duration information is the most direct rhythm information, which can realize rhythm control in the phoneme level. Style information contains an average rhythm of the entire sentence. We perform MFA \cite{mcauliffe2017montreal} on ground true audio to extract duration information of each phoneme. In order to achieve external rhythm control, when calculating the $j$-th alignment weight at current decoding step $i$, $\omega_{i,j}$ represents the probability of focused hidden state remain unchanged while $1-\omega_{i,j}$ represents the probability of focused hidden state move one step forward.
 
 Applying the above two assumptions to basic procedure mentioned in Sec.~\ref{related_work}, mathematical expressions of RC-Attention are formulated as follows:
\begin{equation}
e_{i,j}= Tanh(q_i+h_j )
\label{equbah}
\end{equation}
The energy function is the same as Bahdanau attention \cite{bahdanau2014neural} where $q_i$ is processed hidden state of attention LSTM in the $i$-th step, which contains style information provided by global style tokens and acoustic information.
\begin{equation}
\hat{\omega_{i,j}}=DNN(\,e_{i,j}\,,\,L_{i,j}\,)
\end{equation}
where $L_{i,j}$ is the duration embedding. The shape of extracted duration embedding sequences is [Batch size, Max Length] after padding as part of the input of attention mechanism. Here, we simply ultilize a linear project as the DNN layer.
\begin{equation}
 \omega_{i,j}=\delta \,(\,\hat{\omega_{i,j}}\,)
\end{equation}
where $\delta$ represents sigmoid function which ensures value of $\omega$ is between 0 and 1.

After alignment vector $a_{i,j}$ is calculated as shown in Eq.~(\ref{equ4}), , we apply a trainable scalar $\omega_{i,j}$ in our Rhythm-controllable Attention to control rhythm.
\begin{equation}
 a_{i,j}=(1-\omega_{i,j-1})a_{i-1,j-1}+\omega_{i,j}\, a_{i-1,j}
\label{equ9}
\end{equation}
The bigger the value of $\omega_{i,j}$, the more difficult it is to transfer the weight from the previous encoder hidden state, and the slower rhythm of speech.

\section{Experiments}
\label{experiment}
\subsection{Experimental setup}
To verify effectiveness of the proposed method, we compare widely used alignment methods. This section will introduce training corpus, compared models and implementation details. 

\textbf{Training corpus:} Considered distribution of rhythm, we take an self-built emotional corpus as training corpus in which distribution of duration is more diverse. The training corpus contains <text, audio, emotion label> pairs, covering three different emotion categories (neutral, happy and angry). Each sentence has an average length of 14 words. The whole corpus consists of $7000$ utterances, including $2334$ happy utterances, $2332$ angry utterances and $2334$ neutral utterances. It has a total length of approximately $8.4$ hours uttered by a young female. Audios are sampled at $48$ kHz, but all utterances are downsampled to $22$ kHz and are represented as 80 dimensional mel spectrograms in the experiments.

\textbf{Compared models:}
To evaluate the effects of rhythm-controllable attention, models on which we conduct experiments for comparison include:
\begin{enumerate}
\item Baseline: Location Sensitive Attention \cite{chorowski2015attention}.
\item GMM: GMMv2b \cite{graves2013generating} with five GMM mixture number.
\item FA: Forward Attention \cite{zhang2018forward} with transition agent.
\item Proposed: Proposed model describes in Sec.~\ref{model}
\end{enumerate}

\textbf{Implementation details:}
The parameter settings of text encoder, decoder and style encoder structure are the same as original Tacotron2 \cite{shen2018natural} and Global Style Tokens \cite{wang2018style}, respectively. For style encoder, style embedding channel and number of style tokens is equal to $256$ and ten, respectively. A MelGAN vocoder \cite{DBLP:conf/nips/KumarKBGTSBBC19} is used in our experiments as the vocoder. For GMMv2b Attention, we set GMM mixture number, delta bias and sigma bias to $5$, $0.2$ and $2.0$, respectively.
In our experiment, training corpus is randomly split into a training set ($6900$) and validation set ($100$). 
All models are trained for at least $100k$ steps with a batch size of $32$ using the Adam optimizer \cite{kingma2014adam}. In the inferring phrase, a reference utterance from each of styles is used.

% 225 180pt
\subsection{Robustness evaluation}
To measure robustness of proposed method, each model synthesizes $40$ speech samples for each emotion. Two subjects randomly choose ten samples from $120$ speech synthesized by different models. All texts synthesized are out-of-domain sentences which have an average length of $147$ words. Results are summarized in Table~\ref{tab:Objective Evaluation}. Even though sentences are nearly ten times longer than sentences in training corpus, proposed attention can still achieve the best robustness.

\begin{figure}[htp]
\centering
\includegraphics[scale=0.2]{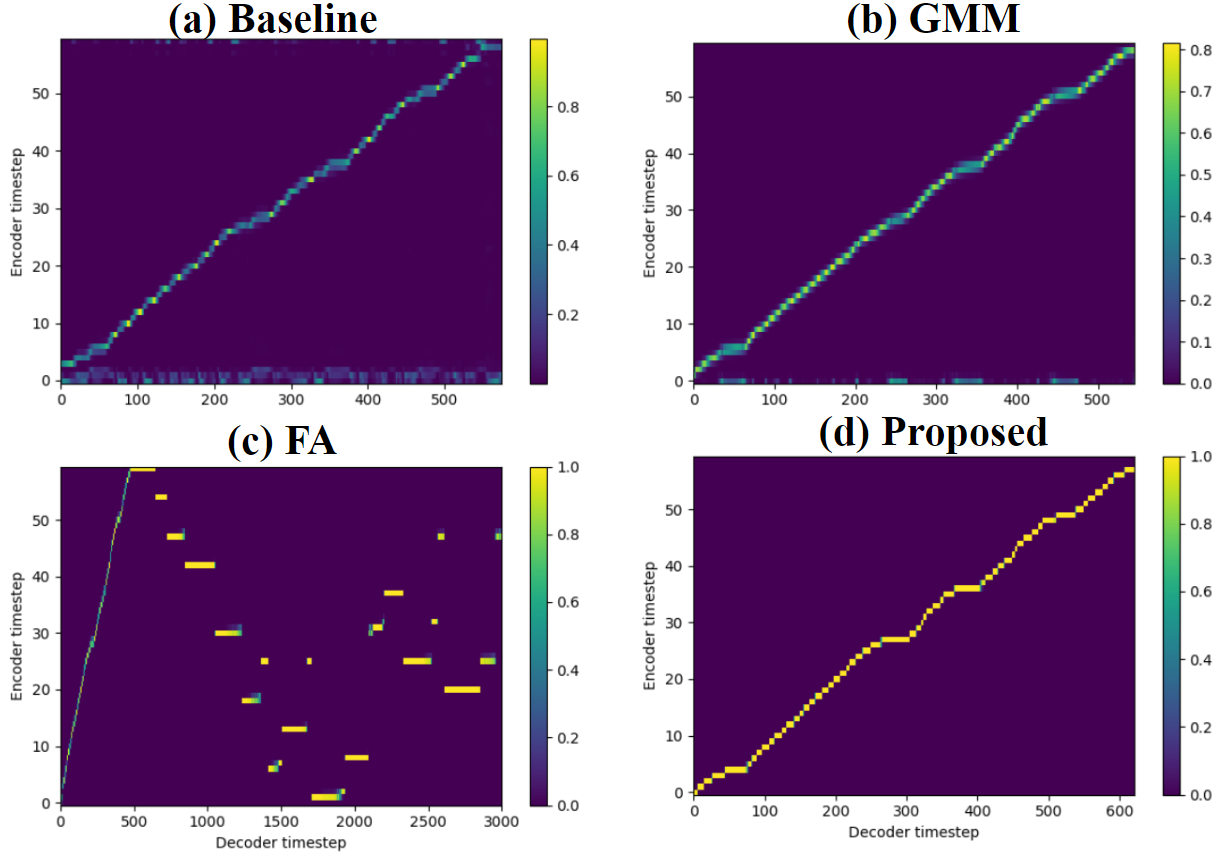}  % 图片路径
\caption{Alignment visualization of different attentions.}  % 图片标题
\label{fig:compared_alignments}    % 标签，用来引用
\end{figure}

Moreover, in order to intuitively demonstrate robustness of different attention mechanisms, we visualize alignment diagrams as shown in Fig.~\ref{fig:compared_alignments}. For baseline, a blurred horizontal line exists at the bottom of alignment diagram, which indicates that baseline would pay attention to the first few encoder hidden states instead of only focusing on one encoder hidden state at each decoding step. Even though alignment of GMMv2b attention shows a more clear diagonal trend than baseline and Forward Attention, there still exist blurred situations associated with pauses, which tends to generate speech with gibberish. Nevertheless, the proposed attention creates a clearer alignment without dispersing weights on unrelated hidden states at each time step.

\begin{table}[t]
  \caption{Number of word errors from selected models. The lower number means the better robustness performance and the bold indicates the best performance in all the models. *(Total 1426 words)* }
  \label{tab:Objective Evaluation}
  \centering
  \begin{tabular}{cccc}
    \toprule
   \textbf{Model}& \textbf{Skipping} &\textbf{Repeating} & \textbf{Collapse}\\
    \midrule
    Baseline                       & $60$   & $10$   & $98$             \\
    GMM                        & $7$     & $\mathbf{0}$     & $5$              \\
    FA                           & $10$      & $4$      & $4$    \\
    Propoesd                     & $\mathbf{6}$       & $\mathbf{0}$    & $\mathbf{3}$              \\

    \bottomrule
  \end{tabular}

\end{table}

\subsection{Rhythm naturalness evaluation}
\label{Rhythm naturalness evaluation}
To figure out which method produces audio with more natural rhythm, subjects are asked to select preferred samples according to overall impression on the naturalness of rhythm. For each model, we randomly generate 60 samples in which number of samples for every emotion is equal. Totally, there are 240 utterances judged by nine subjects in this experiment. 

Result, summarized in Table~\ref{tab:AB preference}, demonstrates that different alignment methods lead to different rhythm performance. Our proposed model performs better than baseline in all three emotions. Although rhythm generated by GMMv2b attention is slightly better than the proposed attention in term of angry, our proposed model can generate more natural utterances in most cases. Among these compared models, only Forward Attention and proposed attention are capable of controlling rhythm. It is noted that $60$ percent of subjects prefer audio synthesized by proposed method rather than Forward Attention, which illustrates our proposed method achieves better control of rhythm. RC-Attention makes use of four kinds of information to study how to control rhythm. Moreover, it indirectly controls rhythm by influencing energy value. By contrast, Forward Attention try to learn the transition agent from only two kinds of information. Besides, it directly operates transition probability when controls rhythm. The latter method increases probability of occurrence of non-robustness problems, meanwhile rhythm of synthesized speech can be damaged.
\begin{table}[t]
  \caption{AB preference test on naturalness, where “N/P” stands for no preference.}
  \label{tab:AB preference}
  \centering
  \setlength{\tabcolsep}{2mm}{
  \scalebox{1}{
      \begin{tabular}{cccccc}
        \toprule
        \textbf{Emotion} & 
        \textbf{Baseline} & 
        \textbf{GMM} &
        \textbf{FA} &
        \textbf{Proposed} &
        \textbf{N/P} \cr
       
        \midrule \midrule
        \multirow{4}{*}{\textbf{neutral}}    
                                & 7\%      & -          &-          & $\mathbf{55}$\%   & 38\% \\
                                & -         & 12\%       & -          & 32\%    & $\mathbf{56}$\% \\
                                & -         & -          & 3\%        & $\mathbf{86}$\%    & 11\% \\

        \midrule
        \multirow{4}{*}{\textbf{happy}}        
                                & 21\%      & -          &-           & $\mathbf{44}$\%   & 35\% \\
                                & -         & 11\%       & -           & $\mathbf{60}$\%    & 29\% \\
                                & -          & -        & 29\%          & 32\%    & $\mathbf{39}$\% \\

        \midrule
       \multirow{4}{*}{\textbf{angry}}       
                                & 8\%      & -          &-           & $\mathbf{54}$\%   & 38\% \\
                                & -         & $\mathbf{41}$\%        & -           & 26\%    & 33\% \\
                                & -         & -          & 13\%       & $\mathbf{62}$\%    & 25\% \\

        \bottomrule
      \end{tabular}
 }}
\end{table}

\begin{table}[t]
  \caption{The results of subjective MOS tests for emotion expression.}
  \label{tab:MOS}
  \centering
  \begin{tabular}{cccc}
    \toprule
   \textbf{Model}& \textbf{Happy} &\textbf{Angry} & \textbf{Avg}\\
    \midrule
    Baseline                       & $3.60$   & $3.68$   & $3.64$             \\
    GMM                        & $3.67$     & $3.71$     & $3.69$              \\
    FA                           & $3.55$      & $3.63$      & $3.59$    \\
    Propoesd                     & $\mathbf{3.70}$       & $\mathbf{3.79}$    & $\mathbf{3.75}$              \\

    \bottomrule
  \end{tabular}
\end{table}

\subsection{Emotion expression evaluation}
Different emotions have different patterns to control rhythm. Good rhythm control can promote emotion expression. Therefore, we evaluate emotional expression by subjective mean opinion score (MOS). Subjects are asked to give a score between 1 to 5 points in terms of emotion expression. Table~\ref{tab:MOS} demonstrates results that RC-Attention produces best results in two emotions. Score of GMMv2b is the closest to RC-Attention, followed by the baseline and Forward Attention. We attribute Forward Attention's performance in emotional expressiveness to its blunt rhythmic control. The results in Section~\ref{Rhythm naturalness evaluation} show that poor control of rhythm in Forward Attention would lead to unnatural rhythm, furthermore, it would damage emotional expressiveness of synthesized speech.
\section{Conclusions}
\label{conclusion}
In this work, we propose a novel attention mechanism, called RC-Attention, which integrates advantages of previous attentions. RC-attention achieves rhythm control in
phoneme level without destroying robustness. We conduct objective and subjective experiments to evaluate robustness and naturalness of rhythm, respectively. Results demonstrate that RC-Attention can enhance robustness of synthesis compared with other attention mechanisms. Furthermore, our attention mechanism achieves rhythm control with more naturalness.
\section{Acknowledgements}
This work is supported by the Open Project Program of the National Laboratory of Pattern Recognition (NLPR) (202200042) and New Talent Project of Beijing University of Posts and Telecommunications (2021RC37) and the research fund of China testing International Co., Ltd. on speech synthesis of Chinese listening text (HT-202011-374).

\clearpage

\bibliographystyle{IEEEtran}
\bibliography{template}

% \begin{thebibliography}{9}
% \bibitem[1]{Davis80-COP}
%   S.\ B.\ Davis and P.\ Mermelstein,
%   ``Comparison of parametric representation for monosyllabic word recognition in continuously spoken sentences,''
%   \textit{IEEE Transactions on Acoustics, Speech and Signal Processing}, vol.~28, no.~4, pp.~357--366, 1980.
% \bibitem[2]{Rabiner89-ATO}
%   L.\ R.\ Rabiner,
%   ``A tutorial on hidden Markov models and selected applications in speech recognition,''
%   \textit{Proceedings of the IEEE}, vol.~77, no.~2, pp.~257-286, 1989.
% \bibitem[3]{Hastie09-TEO}
%   T.\ Hastie, R.\ Tibshirani, and J.\ Friedman,
%   \textit{The Elements of Statistical Learning -- Data Mining, Inference, and Prediction}.
%   New York: Springer, 2009.
% \bibitem[4]{YourName17-XXX}
%   F.\ Lastname1, F.\ Lastname2, and F.\ Lastname3,
%   ``Title of your INTERSPEECH 2022 publication,''
%   in \textit{Interspeech 2022 -- 23\textsuperscript{rd} Annual Conference of the International Speech Communication Association, September 18-22, Incheon, Korea, Proceedings, Proceedings}, 2022, pp.~100--104.
% \end{thebibliography}

\end{document}